\documentclass[11pt,a4paper]{article}
\usepackage[hyperref]{acl2019}
\usepackage{times}
\usepackage{latexsym}

\usepackage{graphicx}
\usepackage{tabularx}
\usepackage{booktabs}
\usepackage{multirow}
\usepackage{varwidth}
\usepackage{pbox}
\usepackage{nidanfloat}

\usepackage{amssymb}
\usepackage{amsmath}
\usepackage{pifont}

\usepackage{url}

\aclfinalcopy 

\title{LIAAD at SemDeep-5 Challenge: Word-in-Context (WiC)}

\author{Daniel Loureiro, Al\'ipio M\'ario Jorge \\
  LIAAD - INESC TEC \\
  Faculty of Sciences - University of Porto, Portugal \\
  {\tt dloureiro@fc.up.pt, amjorge@fc.up.pt}}

\date{}

\begin{document}
\maketitle
\begin{abstract}
  This paper describes the LIAAD system that was ranked second place in the Word-in-Context challenge (WiC) featured in SemDeep-5.
  Our solution is based on a novel system for Word Sense Disambiguation (WSD) using contextual embeddings and full-inventory sense embeddings. We adapt this WSD system, in a straightforward manner, for the present task of detecting whether the same sense occurs in a pair of sentences.
  Additionally, we show that our solution is able to achieve competitive performance even without using the provided training or development sets, mitigating potential concerns related to task overfitting.
\end{abstract}

\section{Task Overview}

The Word-in-Context (WiC) \cite{wic2019naacl} task aims to evaluate the ability of word embedding models to accurately represent context-sensitive words. In particular, it focuses on polysemous words which have been hard to represent as embeddings due to the meaning conflation deficiency \cite{CamachoCollados2018FromWT}. The task's objective is to detect if target words occurring in a pair of sentences carry the same meaning.

Recently, contextual word embeddings from ELMo \cite{peters2018deep} or BERT \cite{bert_naacl} have emerged as the successors to traditional embeddings. With this development, word embeddings have become context-sensitive by design and thus more suitable for representing polysemous words. However, as shown by the experiments of \cite{wic2019naacl}, they are still insufficient by themselves to reliably detect meaning shifts.

In this work, we propose a system designed for the larger task of Word Sense Disambiguation (WSD), where words are matched with specific senses, that can detect meaning shifts without being trained explicitly to do so. Our WSD system uses contextual word embeddings to produce sense embeddings, and has full-coverage of all senses present in WordNet 3.0 \cite{Fellbaum2000WordNetA}. In \citet{lmmsacl2019} we provide more details about this WSD system, called LMMS (Language Modelling Makes Sense), and demonstrate that it's currently state-of-the-art for WSD. For this challenge, we employ LMMS in two straightforward approaches: checking if the disambiguated senses are equal, and training a classifier based on the embedding similarities. Both approaches perform competitively, with the latter taking the second position in the challenge ranking, and the former trailing close behind even though it's tested directly on the challenge, forgoing the training and development sets.

\section{System Description}

LMMS has two useful properties: 1) uses contextual word embeddings to produce sense embeddings, and 2) covers a large set of over 117K senses from WordNet 3.0. The first property allows for comparing precomputed sense embeddings against contextual word embeddings generated at test-time (using the same language model). The second property makes the comparisons more meaningful by having a large selection of senses at disposal for comparison.

\subsection{Sense Embeddings}

Given the meaning conflation deficiency issue with traditional word embeddings, several works have focused on adapting Neural Language Models (NLMs) to produce word embeddings that are more sense-specific. In this work, we start producing sense embeddings from the approach used by recent works in contextual word embeddings, particularly context2vec \cite{Melamud2016context2vecLG} and ELMo \cite{peters2018deep}, and introduce some improvements towards full-coverage and more accurate representations.

\subsubsection{Using Supervision}

Our set of full-coverage WordNet sense embeddings is bootstrapped from the SemCor corpus \cite{Miller1994UsingAS}. Sentences containing sense-annotated tokens (or spans) are processed by a NLM in order to obtain contextual embeddings for those tokens. After collecting all sense-labeled contextual embeddings, each sense embedding ($\vec{v}_s$) is determined by averaging its corresponding contextual embeddings. Formally, given $n$ contextual embeddings $\vec{c}$ for some sense $s$:

$$\vec{v}_{s} = \frac{1}{n}\sum_{i=1}^{n}\vec{c}_i$$

In this work, we used BERT as our NLM. For replicability, these are the relevant details: 1024 embedding dimensions, 340M parameters, cased. Embeddings result from the sum of top 4 layers ([-1, -4]). Moreover, since BERT uses WordPiece tokenization that doesn't always map to token-level annotations, we use the average of subtoken embeddings as the token-level embedding.

\subsubsection{Extending Supervision}

Despite its age, SemCor is still the largest sense-annotated corpus. The lack of larger sets of sense annotations is a major limitation of supervised approaches for WSD \cite{Le2018ADD}. We address this issue by taking advantage of the semantic relations in WordNet to extend the annotated signal to other senses.
Missing sense embeddings are inferred (i.e. imputed) from the aggregation of sense embeddings at different levels of abstraction from WordNet's ontology. Thus, a synset embedding corresponds to the average of all of its sense embeddings, a hypernym embedding corresponds to the average of all of its synset embeddings, and a lexname embedding corresponds to the average of a larger set of synset embeddings. All lower abstraction representations are created before next-level abstractions to ensure that higher abstractions make use of lower-level generalizations. More formally, given all missing senses in WordNet $\hat{s} \in {W}$, their synset-specific sense embeddings $S_{\hat{s}}$, hypernym-specific synset embeddings $H_{\hat{s}}$, and lexname-specific synset embeddings $L_{\hat{s}}$, the procedure has the following stages:
$$
\begin{matrix}
(1) & if |S_{\hat{s}}| > 0 , & \vec{v}_{\hat{s}} = \frac{1}{|S_{\hat{s}}|}\sum\vec{v}_{s} , \forall \vec{v}_{s} \in S_{\hat{s}} \\\\

(2) & if |H_{\hat{s}}| > 0 , & \vec{v}_{\hat{s}} = \frac{1}{|H_{\hat{s}}|}\sum\vec{v}_{syn} , \forall \vec{v}_{syn} \in H_{\hat{s}} \\\\

(3) & if |L_{\hat{s}}| > 0 , & \vec{v}_{\hat{s}} = \frac{1}{|L_{\hat{s}}|}\sum\vec{v}_{syn} , \forall \vec{v}_{syn} \in L_{\hat{s}}
\end{matrix}
$$

\subsubsection{Leveraging Glosses}

There's a long tradition of using glosses for WSD, perhaps starting with the popular work of \citet{Lesk1986AutomaticSD}. As a sequence of words, the information contained in glosses can be easily represented in semantic spaces through approaches used for generating sentence embeddings. While there are many methods for generating sentence embeddings, it's been shown that a simple weighted average of word embeddings performs well \cite{Arora2017ASB}.

Our contextual embeddings are produced from NLMs that employ attention mechanisms, assigning more importance to some tokens over others. As such, these embeddings already come `pre-weighted' and we embed glosses simply as the average of all of their contextual embeddings (without preprocessing). We've found that introducing synset lemmas alongside the words in the gloss helps induce better contextualized embeddings (specially when glosses are short). Finally, we make our dictionary embeddings ($\vec{v}_d$) sense-specific, rather than synset-specific, by repeating the lemma that's specific to the sense alongside all of the synset's lemmas and gloss words. The result is a sense-level embedding that is represented in the same space as the embeddings we described in the previous section, and can be trivially combined through concatenation (previously $L_2$ normalized).

Given that both representations are based on the same NLM, we can make predictions for contextual embeddings of target words $w$ (again, using the same NLM) at test-time by simply duplicating those embeddings, aligning contextual features against sense and dictionary features when computing cosine similarity. Thus, we have sense embeddings $\vec{v}_s$, to be matched against duplicated contextual embeddings $\vec{c}_w$, represented as follows:

$$
\vec{v}_s = \begin{bmatrix}
  ||\vec{v}_s||_2\\
  ||\vec{v}_d||_2
  \end{bmatrix},
\vec{c}_w = \begin{bmatrix}
  ||\vec{c}_w||_2\\
  ||\vec{c}_w||_2
  \end{bmatrix}
$$

\subsection{Sense Disambiguation}

\begin{figure}[htb]
  \centering
  \includegraphics[width=0.35\textwidth]{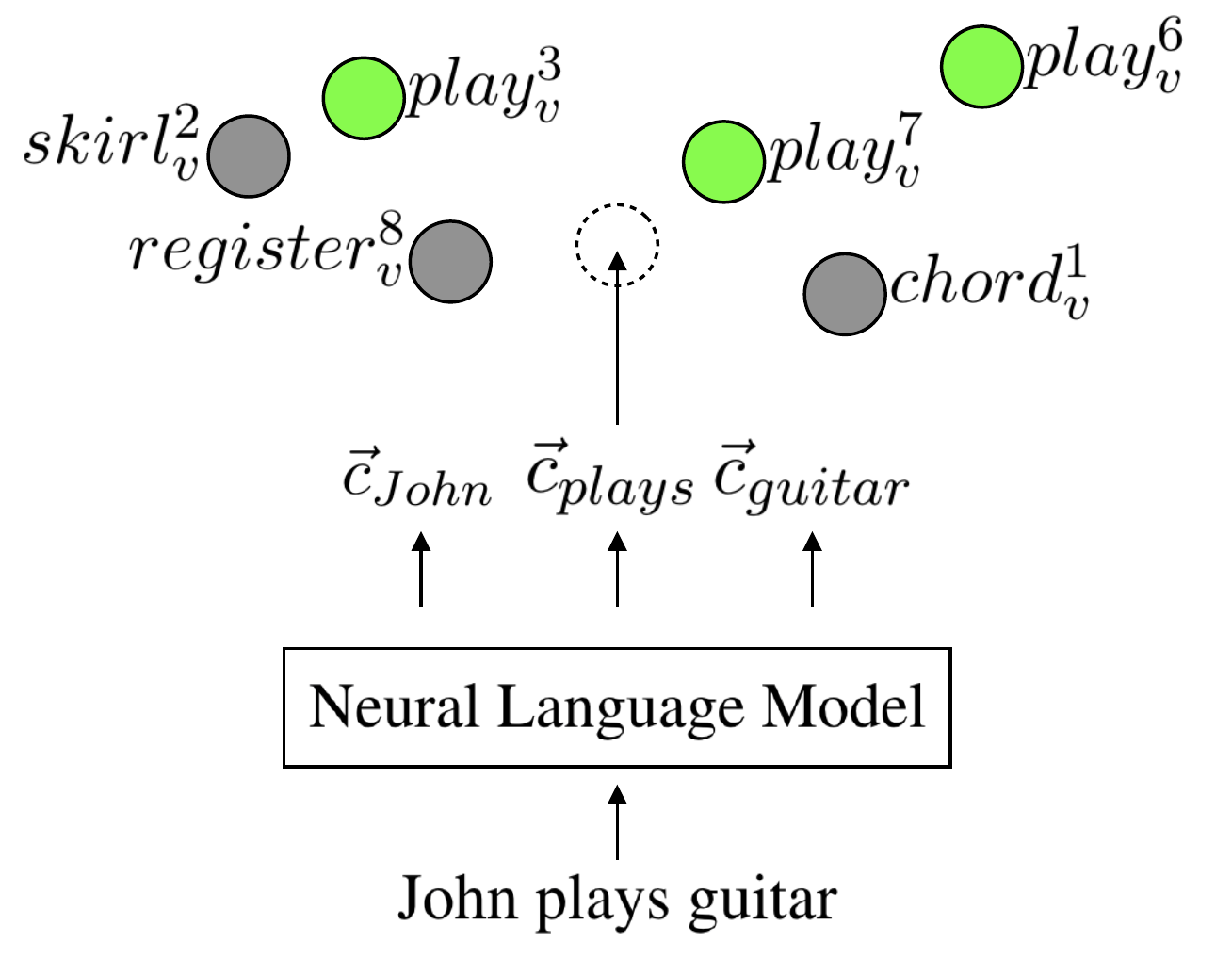}
  \caption{Illustration of our $k$-NN approach for WSD, which relies on full-coverage sense embeddings represented in the same space as contextualized embeddings.}
  \label{fig:knn}
\end{figure}

Having produced our set of full-coverage sense embeddings, we perform WSD using a simple Nearest-Neighbors ($k$-NN) approach, similarly to \citet{Melamud2016context2vecLG} and \citet{peters2018deep}. We match the contextual word embedding of a target word against the sense embeddings that share the word's lemma (see Figure \ref{fig:knn}). Matching is performed using cosine similarity (with duplicated features on the contextual embedding for alignment, as explained in 2.1.3), and the top match is used as the disambiguated sense.

\subsection{Binary Classification}

The WiC task calls for a binary judgement on whether the meaning of a target word occurring in a pair of sentences is the same or not. As such, our most immediate solution is to perform WSD and base our decision on the resulting senses. This approach performs competitively, but we've still found it worthwhile to use WiC's data to train a classifier based on the strengths of similarities between contextual and sense embeddings. In this section we explore the details of both approaches.

\begin{figure*}[b]
  \centering
  \includegraphics[width=\textwidth]{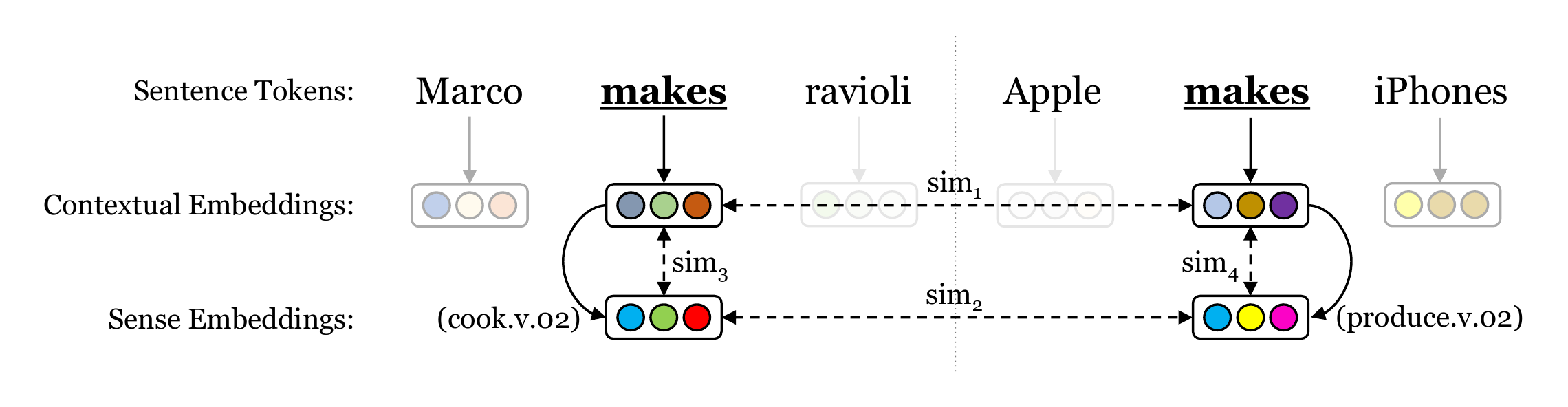}
  \caption{Components and interactions involved in our approaches. The sim$_n$ labels correspond to cosine similarities between the related embeddings. Sense embeddings obtained from 1-NN matches of contextual embeddings.}
  \label{fig:sims}
\end{figure*}

\subsubsection{Sense Comparison}

Our first approach is a straightforward comparison of the disambiguated senses assigned to the target word in each sentence. Considering the example in Figure \ref{fig:sims}, this approach simply requires checking if the sense $cook_{v}^2$ assigned to `makes' in the first sentence equals the sense $produce_{v}^2$ assigned to the same word in the second sentence.

\subsubsection{Classifying Similarities}

The WSD procedure we describe in this paper represents sense embeddings in the same space as contextual word embeddings. Our second approach exploits this property by considering the similarities (including between different embedding types) that can be seen in Figure \ref{fig:sims}. In this approach, we take advantage of WiC's training set to learn a Logistic Regression Binary Classifier based on different sets of similarities. The choice of Logistic Regression is due to its explainability and lightweight training, besides competitive performance. We use sklearn's implementation (v0.20.1), with default parameters.

\section{Results}

The best system we submitted during the evaluation period of the challenge was a Logistic Regression classifier trained on two similarity features (sim$_1$ and sim$_2$, or contextual and sense-level). We obtained slightly better results with a classifier trained on all four similarities shown in Figure \ref{fig:sims}, but were unable to submit that system due to the limit of a maximum of three submissions during evaluation. Interestingly, the  simple approach described in 2.3.1 achieved a competitive performance of 66.3 accuracy, without being trained or fine-tuned on WiC's data. Performance of best entries and baselines can be seen on Table \ref{tab:rank}.

\begin{table}[]
\centering
\begin{tabular}{|c|c|}
\hline
\textbf{Submission}                                           & \textbf{Acc.} \\ \hline
\begin{tabular}[c]{@{}c@{}}SuperGlue\\\cite{DBLP:journals/corr/abs-1905-00537}\end{tabular}     & 68.36         \\ \hline
\begin{tabular}[c]{@{}c@{}}LMMS\\(Ours)\end{tabular}         & 67.71         \\ \hline
\begin{tabular}[c]{@{}c@{}}Ensemble\\\cite{GariWiC}\end{tabular}      & 66.71         \\ \hline
\begin{tabular}[c]{@{}c@{}}ELMo-weighted\\\cite{AnsellWiC}\end{tabular} & 61.21         \\ \hline\hline
BERT-large                                                    & 65.5          \\ \hline
Context2vec                                                   & 59.3          \\ \hline
ELMo-3                                                        & 56.5          \\ \hline
Random                                                        & 50.0          \\ \hline
\end{tabular}
\caption{Challenge results at the end of the evaluation period. Bottom results correspond to baselines.}
\label{tab:rank}
\end{table}

\section{Analysis}

In this section we provide additional insights regarding our best approach. In Table \ref{tab:models}, we show how task performance varies with the similarities considered.

\begin{table}[h]
\centering
\begin{tabular}{|c|r|c|c|}
\hline
\textbf{Model} & \multicolumn{1}{c|}{\textbf{sim$_n$}} & \textbf{Dev} & \textbf{Test}  \\ \hline
M0             & \multicolumn{1}{c|}{N/A}            & 68.18        & 66.29          \\ \hline
M1             & 1                                   & 67.08        & 64.64          \\ \hline
M2             & 2                                   & 66.93        & 66.21          \\ \hline
M3             & 1, 2                                & 68.50        & 67.71    \\ \hline
M4             & 1, 2, 3, 4                          & 69.12        & \textbf{68.07} \\ \hline
\end{tabular}
\caption{Accuracy of our different models. M0 wasn't trained on WiC data, the other models were trained on different sets of similarites. We submitted M3, but achieved slightly improved results with M4.}
\label{tab:models}
\end{table}

\begin{figure}[htb]
  \centering
  \includegraphics[width=0.45\textwidth]{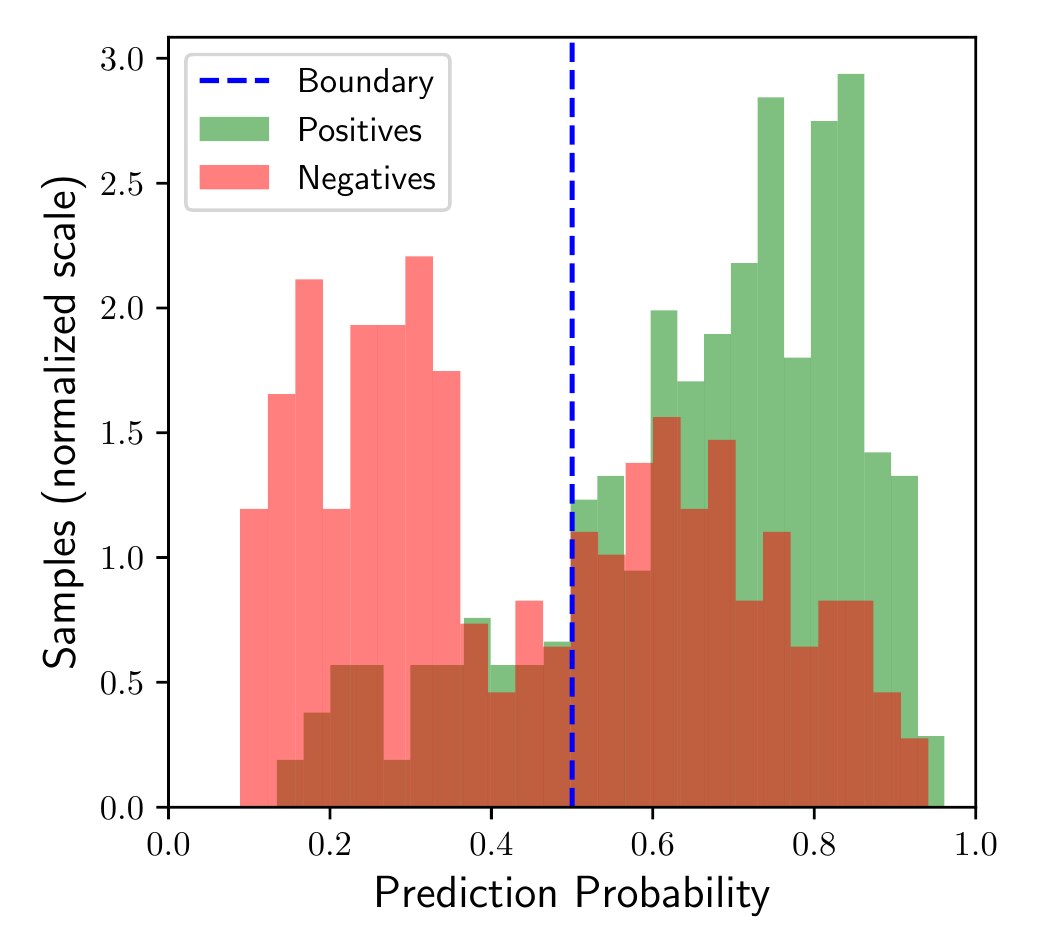}
  \caption{Distribution of Prediction Probabilities across labels, as evaluated by our best model on the development set.}
  \label{fig:prob}
\end{figure}

\begin{figure}[!htb]
  \centering
  \includegraphics[width=0.45\textwidth]{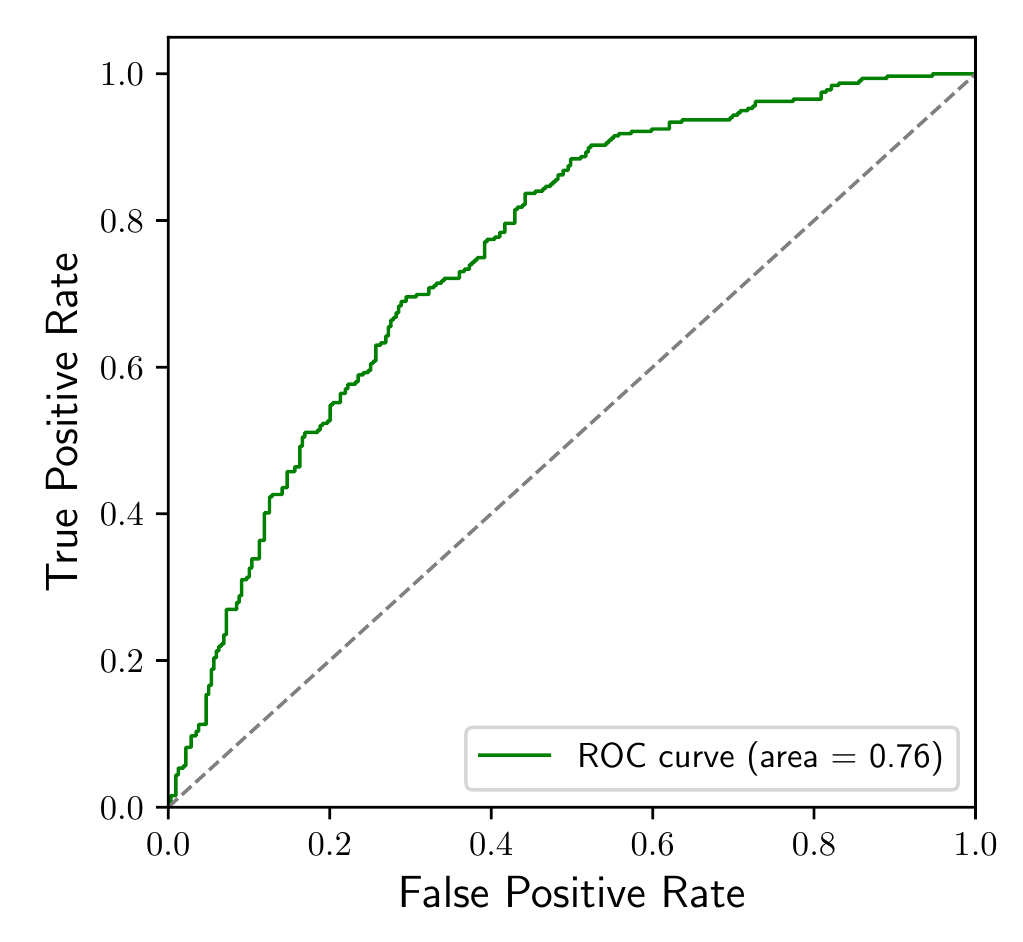}
  \caption{ROC curve for results of our best model on the development set.}
  \label{fig:curve}
\end{figure}

We determined that our best system (M4, using four features) obtains a precision of 0.65, recall of 0.82, and F1 of 0.73 on the development set, showing a relatively high proportion of false positives (21.6\% vs. 9.25\% of false negatives). This skewness can also be seen in the probability distribution chart at Figure \ref{fig:prob}. Additionally, we also present a ROC curve for this system at Figure \ref{fig:curve} for a more detailed analysis of the system's performance.

\section{Conclusion and Future Work}

We've found that the WiC task can be adequately solved by systems trained for the larger task of WSD, specially if they're based on contextual embeddings, and when compared to the reported baselines. Still, we've found that the WiC dataset can be useful to learn a classifier that builds on top of the WSD system for improved performance on WiC's task of detecting shifts in meaning. In future work, we believe this improved ability to detect shifts in meaning can also assist WSD, particularly in generating semi-supervised datasets. We share our code and data at \href{https://github.com/danlou/lmms}{github.com/danlou/lmms}.

\section*{Acknowledgements}

This work is financed by National Funds through the Portuguese funding agency, FCT - Funda\c{c}\~{a}o para a Ci\^{e}ncia e a Tecnologia within project: UID/EEA/50014/2019.

\bibliography{acl2019}

\begin{thebibliography}{14}
\expandafter\ifx\csname natexlab\endcsname\relax\def\natexlab#1{#1}\fi

\bibitem[{Ansell et~al.(2019)Ansell, Bravo-Marquez, and Pfahringer}]{AnsellWiC}
Alan Ansell, Felipe Bravo-Marquez, and Bernhard Pfahringer. 2019.
\newblock An elmo-inspired approach to semdeep-5's word-in-context task.
\newblock In \emph{SemDeep-5@IJCAI 2019}, page forthcoming.

\bibitem[{Arora et~al.(2017)Arora, Liang, and Ma}]{Arora2017ASB}
Sanjeev Arora, Yingyu Liang, and Tengyu Ma. 2017.
\newblock \href {https://openreview.net/forum?id=SyK00v5xx} {A simple but
  tough-to-beat baseline for sentence embeddings}.
\newblock In \emph{International Conference on Learning Representations
  (ICLR)}.

\bibitem[{Camacho-Collados and Pilehvar(2018)}]{CamachoCollados2018FromWT}
Jose Camacho-Collados and Mohammad~Taher Pilehvar. 2018.
\newblock \href {https://doi.org/10.1613/jair.1.11259} {From word to sense
  embeddings: A survey on vector representations of meaning}.
\newblock \emph{J. Artif. Int. Res.}, 63(1):743--788.

\bibitem[{Devlin et~al.(2019)Devlin, Chang, Lee, and Toutanova}]{bert_naacl}
Jacob Devlin, Ming-Wei Chang, Kenton Lee, and Kristina Toutanova. 2019.
\newblock \href {https://www.aclweb.org/anthology/N19-1423} {{BERT}:
  Pre-training of deep bidirectional transformers for language understanding}.
\newblock In \emph{Proceedings of the 2019 Conference of the North {A}merican
  Chapter of the Association for Computational Linguistics: Human Language
  Technologies, Volume 1 (Long and Short Papers)}, pages 4171--4186,
  Minneapolis, Minnesota. Association for Computational Linguistics.

\bibitem[{Fellbaum(1998)}]{Fellbaum2000WordNetA}
Christiane Fellbaum. 1998.
\newblock In \emph{WordNet : an electronic lexical database}. MIT Press.

\bibitem[{Le et~al.(2018)Le, Postma, Urbani, and Vossen}]{Le2018ADD}
Minh Le, Marten Postma, Jacopo Urbani, and Piek Vossen. 2018.
\newblock \href {https://www.aclweb.org/anthology/C18-1030} {A deep dive into
  word sense disambiguation with {LSTM}}.
\newblock In \emph{Proceedings of the 27th International Conference on
  Computational Linguistics}, pages 354--365, Santa Fe, New Mexico, USA.
  Association for Computational Linguistics.

\bibitem[{Lesk(1986)}]{Lesk1986AutomaticSD}
Michael Lesk. 1986.
\newblock \href {https://doi.org/10.1145/318723.318728} {Automatic sense
  disambiguation using machine readable dictionaries: How to tell a pine cone
  from an ice cream cone}.
\newblock In \emph{Proceedings of the 5th Annual International Conference on
  Systems Documentation}, SIGDOC '86, pages 24--26, New York, NY, USA. ACM.

\bibitem[{Loureiro and Jorge(2019)}]{lmmsacl2019}
Daniel Loureiro and Al{\'\i}pio Jorge. 2019.
\newblock Language modelling makes sense: Propagating representations through
  wordnet for full-coverage word sense disambiguation.
\newblock In \emph{Proceedings of the 57th Annual Meeting of the Association
  for Computational Linguistics}, page forthcoming, Florence, Italy.
  Association for Computational Linguistics.

\bibitem[{Melamud et~al.(2016)Melamud, Goldberger, and
  Dagan}]{Melamud2016context2vecLG}
Oren Melamud, Jacob Goldberger, and Ido Dagan. 2016.
\newblock \href {https://doi.org/10.18653/v1/K16-1006} {context2vec: Learning
  generic context embedding with bidirectional {LSTM}}.
\newblock In \emph{Proceedings of The 20th {SIGNLL} Conference on Computational
  Natural Language Learning}, pages 51--61, Berlin, Germany. Association for
  Computational Linguistics.

\bibitem[{Miller et~al.(1994)Miller, Chodorow, Landes, Leacock, and
  Thomas}]{Miller1994UsingAS}
George~A. Miller, Martin Chodorow, Shari Landes, Claudia Leacock, and Robert~G.
  Thomas. 1994.
\newblock \href {https://www.aclweb.org/anthology/H94-1046} {Using a semantic
  concordance for sense identification}.
\newblock In \emph{{HUMAN} {LANGUAGE} {TECHNOLOGY}: Proceedings of a Workshop
  held at Plainsboro, New Jersey, March 8-11, 1994}.

\bibitem[{Peters et~al.(2018)Peters, Neumann, Iyyer, Gardner, Clark, Lee, and
  Zettlemoyer}]{peters2018deep}
Matthew Peters, Mark Neumann, Mohit Iyyer, Matt Gardner, Christopher Clark,
  Kenton Lee, and Luke Zettlemoyer. 2018.
\newblock \href {https://doi.org/10.18653/v1/N18-1202} {Deep contextualized
  word representations}.
\newblock In \emph{Proceedings of the 2018 Conference of the North {A}merican
  Chapter of the Association for Computational Linguistics: Human Language
  Technologies, Volume 1 (Long Papers)}, pages 2227--2237, New Orleans,
  Louisiana. Association for Computational Linguistics.

\bibitem[{Pilehvar and Camacho-Collados(2019)}]{wic2019naacl}
Mohammad~Taher Pilehvar and Jose Camacho-Collados. 2019.
\newblock Wic: the word-in-context dataset for evaluating context-sensitive
  meaning representations.
\newblock In \emph{Proceedings of NAACL}, Minneapolis, United States.

\bibitem[{Soler et~al.(2019)Soler, Apidianaki, and Allauzen}]{GariWiC}
Aina~Gar{\'\i} Soler, Marianna Apidianaki, and Alexandre Allauzen. 2019.
\newblock Limsi-multisem at the ijcai semdeep-5 wic challenge: Context
  representations for word usage similarity estimation.
\newblock In \emph{SemDeep-5@IJCAI 2019}, page forthcoming.

\bibitem[{Wang et~al.(2019)Wang, Pruksachatkun, Nangia, Singh, Michael, Hill,
  Levy, and Bowman}]{DBLP:journals/corr/abs-1905-00537}
Alex Wang, Yada Pruksachatkun, Nikita Nangia, Amanpreet Singh, Julian Michael,
  Felix Hill, Omer Levy, and Samuel~R. Bowman. 2019.
\newblock \href {http://arxiv.org/abs/1905.00537} {Superglue: {A} stickier
  benchmark for general-purpose language understanding systems}.
\newblock \emph{CoRR}, abs/1905.00537.

\end{thebibliography}
\bibliographystyle{acl_natbib}

\end{document}